\documentclass[conference]{IEEEtran}

\usepackage{multirow}
\ifCLASSINFOpdf
   \usepackage[pdftex]{graphicx}
   
   \graphicspath{{./pdf/}{./jpeg/}}
   \DeclareGraphicsExtensions{.pdf,.jpeg,.png}
\else
   \usepackage[dvips]{graphicx}
   \graphicspath{{./eps/}}
   \DeclareGraphicsExtensions{.eps}
\fi
\usepackage{mathtools}
\usepackage{color}
\usepackage{algorithm}
\usepackage{algorithmic}

\usepackage{times}
\usepackage{epsfig}
\usepackage{graphicx}
\usepackage{amsmath}
\usepackage{amssymb}
\usepackage{romannum}
\usepackage{color}
\newcommand{\etal}{\textit{et al}.}

\usepackage{url}

\begin{document}
%
\title{Hierarchical Transfer Convolutional Neural Networks for Image Classification\\[.75ex] 
  {\normalfont\large 
     Xishuang Dong, Hsiang-Huang Wu, Yuzhong Yan, Lijun Qian%
  }\\[-1.5ex]
 }

\author{ 
\authorblockA{Center of Excellence in Research and Education for Big Military Data Intelligence (CREDIT) \\
Prairie View A\&M University, Texas A\&M University System  \\
Prairie View, TX 77446, USA \\
Email: dongxishuang@gmail.com, virtuoso.wu@gmail.com, yyuzhong@gmail.com, liqian@pvamu.edu }

}


%


\maketitle

\begin{abstract}
In this paper, we address the issue of how to enhance the generalization performance of convolutional neural networks (CNN) in the early learning stage for image classification. This is motivated by real-time applications that require the generalization performance of CNN to be  satisfactory within limited training time. In order to achieve this, a novel hierarchical transfer CNN framework is proposed. It consists of a group of shallow CNNs and a cloud CNN, where the shallow CNNs are trained firstly and then the first layers of the trained shallow CNNs are used to initialize the first layer of the cloud CNN. This method will boost the generalization performance of the cloud CNN significantly,  especially during the early stage of training. Experiments using CIFAR-10 and ImageNet datasets are performed to examine the proposed method. Results demonstrate the improvement of testing accuracy is $12\%$ on average and as much as $20\%$  for the CIFAR-10 case while $5\%$ testing accuracy improvement for the ImageNet case during the early stage of learning. It is also shown that universal improvements of testing accuracy are obtained across different settings of dropout and number of shallow CNNs. 
\end{abstract}

\begin{IEEEkeywords} Convolutional Neural Networks;Transfer Deep Learning; Image Classification;  \end {IEEEkeywords}

%
\IEEEpeerreviewmaketitle

\section{Introduction}
\label{sec1}


Convolutional neural networks (CNN) has significantly promoted developments of visual processing tasks such as image classification \cite{NIPS2012_0534, he2016deep}, object detection \cite{ren2015faster} and tracking \cite{Karpathy2014}, and semantic segmentation \cite{long2015fully}, which benefits from accessible big datasets such as ImageNet \cite{deng2009imagenet} and YouTube-BoundingBoxes \cite{real2017youtube} that can be used to train large-scale models. CNN based state-of-the-art deep learning architectures such as AlexNet \cite{NIPS2012_0534}, VGG \cite{simonyan2014very}, and GoogleNet \cite{szegedy2015going} are proposed to make rapid progress in image classification. Millions of annotated samples in these big datasets help us to estimate appropriate parameters in these architectures successfully. Further work has advanced CNN by combining CNN with other deep learning models. For example, Wang \etal \cite{wang2016cnn} combine CNN with recurrent neural networks (RNN) for multi-label image classification. In addition, combining  CNN  with  autoencoder  \cite{yim2015rotating, yang2015weakly} has been verified to solve tasks such as face rotation and intrinsic transformations for objects.

\begin{figure}[!htb]
  \centering
  \includegraphics[width= 0.5\textwidth]{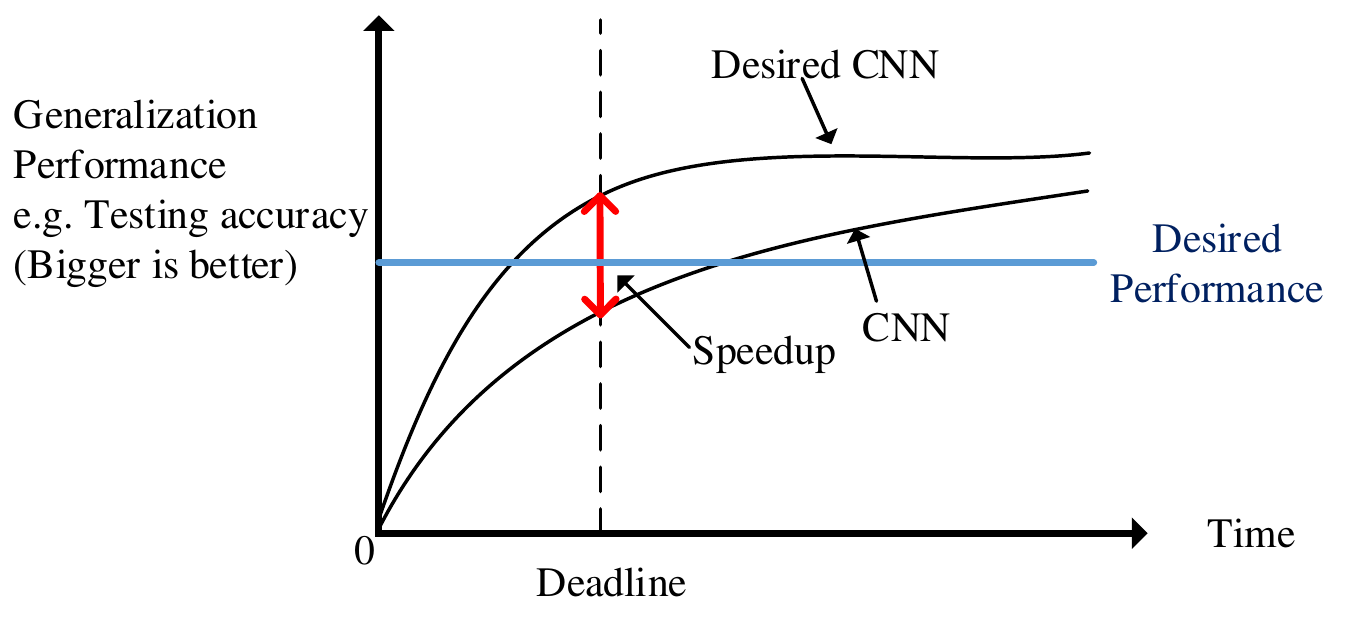}\\
  \caption{Requirements of CNN based real-time applications. A desired generalization performance of CNN may be required by the application while the training time is limited (up to a given deadline). }\label{fig:rt}
\end{figure}

\begin{figure*}[!htb]
  \centering
  \includegraphics[width= 0.95\textwidth]{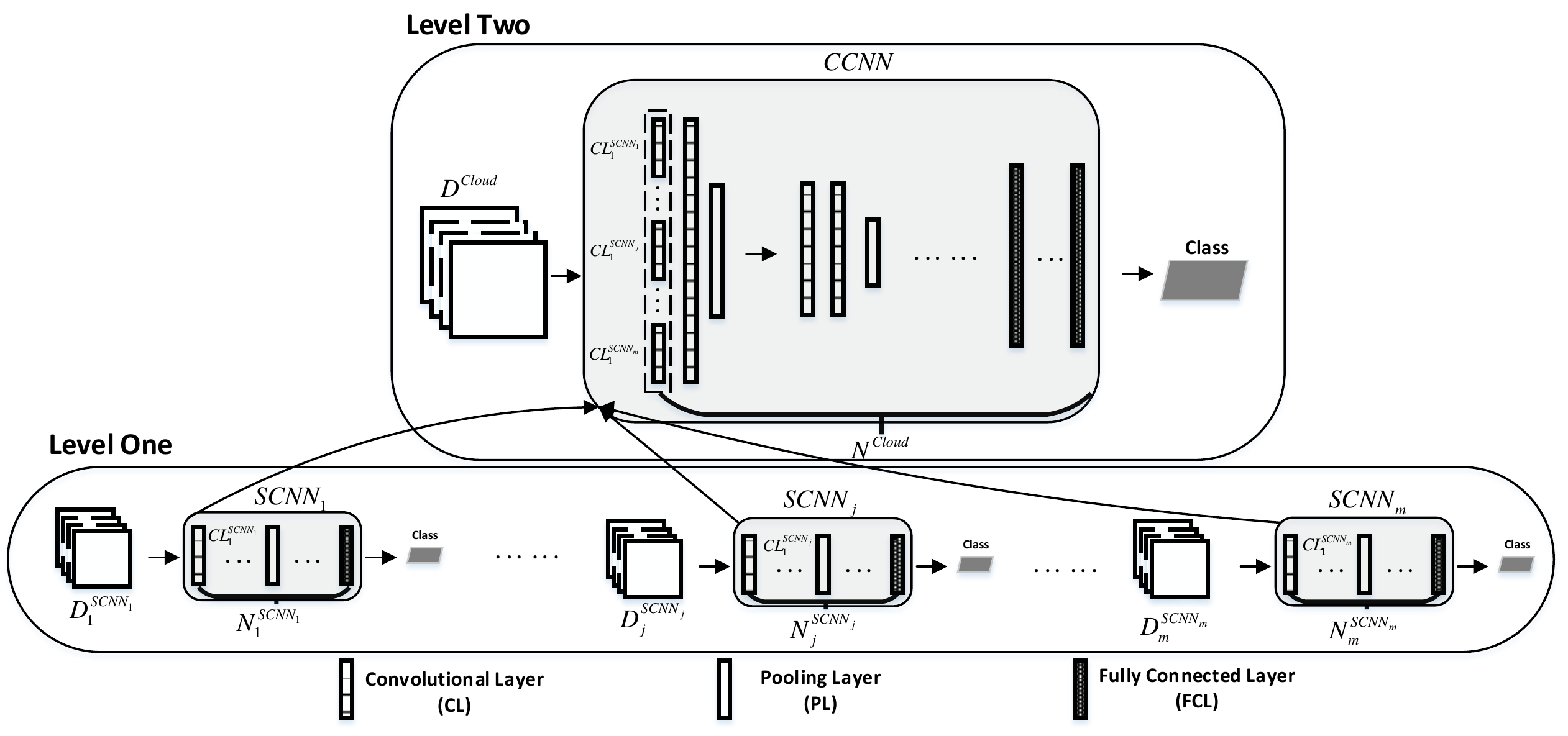}\\
  \caption{Architecture of hierarchical transfer convolutional neural networks (HTCNN), where $SCNN_{j}$ denotes the jth shallow CNN, $CCNN$ means the cloud CNN, $D_{j}$ is training data for building $SCNN_{j}$, and $D^{cloud}$ is training data for constructing $CCNN$. $CL_{1}^{SCNN_{j}}$ denotes the first convolutional layer  of $SCNN_{j}$. $N^{SCNN_{j}}$ and $N^{cloud}$ are the number of layers including convolutional layers, pooling layers, and fully connected layers in $SCNN_{j}$ and $CCNN$, respectively. We have $m$ $SCNNs$ in level one.}\label{fig:TTCNN}
\end{figure*}

Despite the encouraging progress of visual processing via CNN, training a large CNN is still too time consuming to meet the deadline in real-time applications. As illustrated in Figure~\ref{fig:rt}, a novel CNN model is needed to speedup the training process in order to meet the deadline. This motivated us to explore novel design of a deep CNN, where a hierarchical transfer CNN (HTCNN) architecture is proposed to enhance CNN generalization in the early learning stage by transferring multi-source of knowledge. As shown in Figure \ref{fig:TTCNN} , the proposed architecture is composed of a group of shallow CNNs and a cloud CNN, where the number of layers of shallow CNNs is less than that of the cloud CNN. It is composed of four steps to implement the hierarchical transfer CNN: (\romannum{1}) designing architectures of the shallow CNNs and the cloud CNN; (\romannum{2}) training the shallow CNNs independently on their own datasets; (\romannum{3}) extracting the first layers of the trained shallow CNNs to initialize the first layer of the cloud CNN; (\romannum{4}) training the cloud CNN including fine-tuning the initialized first layer on a big dataset. 

In summary, our contributions are as follows:

\begin{itemize}

\item We propose a novel scheme of hierarchical transfer CNN (HTCNN) to enhance the generalization performance of CNN in the early learning stage for image classification.  This is important for real-time applications that require the generalization performance of CNN to be  satisfactory within limited training time.

\item The proposed model can merge knowledge from different data sources in a scalable manner to realize transfer CNN.

\item We validate our method by testing on the CIFAR-10 and ImageNet datasets. The results demonstrate the proposed hierarchical transfer CNNs speedup the training significantly with the improvement of testing accuracy of $12\%$ on average and as much as $20\%$ for the CIFAR-10 case and $5\%$ accuracy improvement for the ImageNet case during the early stage of learning.

\end{itemize}

\section{Related Work}
\label{sec3}

Our work has connections with developing novel CNN architectures, and transfer CNN \cite{yosinski2014transferable} that is to extend shallow parts of pre-trained CNNs to complete the target task by fine-tuning the extended CNNs on the target datasets.

\textbf{CNN Architectures.} In recent years, sophisticated architectures of CNN such as GoogleNet \cite{deng2009imagenet} and AlexNet \cite{NIPS2012_0534}   achieve significant progresses in image processing. Further works have pushed advances along two directions: Firstly, developing general architectures of CNN such as ResNet \cite{he2016deep} and FCN \cite{long2015fully} promotes improvement of image classification \cite{targ2016resnet, szegedy2017inception}, and semantic segmentation. Secondly, it is to design novel architectures of CNN for improving performance of special tasks. For instant, U-Net \cite{ronneberger2015u} and V-Net \cite{milletari2016v} are proposed to accomplish medical image processing. Generally, these architectures contain a cloud CNN, where  training the cloud CNN needs hours or even days. 
In addition to the cloud CNN, our model introduces a group of shallow CNNs that are trained independently on datasets, where the architecture of the cloud CNN is deeper than those of the shallow CNNs. 

\textbf{Transfer CNN.} 
Transfer learning \cite{pan2010survey} is to utilize knowledge gained from source domain to improve model performance in the target domain. Transfer CNN attracts extensive attentions and achieves great success in different tasks such as image recognition, object detection, and semantic segmentation \cite{ahmed2008training, oquab2014learning, girshick2014rich, shin2016deep}. Yosinski \etal \cite{yosinski2014transferable} present that the first-layer features would not be specific to a particular dataset or task, while the last layer depends greatly on the specific dataset and task, by experimentally quantifying the generality versus specificity of features conducted with layers of a deep CNN. Most of recent approaches based on transfer CNN are to extract shallow layers or the whole pre-trained models, extend the extracted components by adding extra layers, and train the extended model by fine-tuning or frozen the extracted components for the target tasks \cite{sharif2014cnn, girshick2016region, ng2015deep, huh2016makes, bansal2016pixelnet, shin2016deep, lee2017transfer}. Encouraging performance has been obtained where these pre-trained models are learned from a single data source. Our proposed model is to transfer multiple pre-trained models from multi-source data, which is to use multi-source knowledge to enhance the CNN generalization and speedup the training process.

\section{Model}
\label{sec4}

\textbf{Overview.} Our goal is to design a novel CNN model for image classification to address the challenge of speeding up the training such that the models with high generalization performance can be obtained as early as possible to potentially meet the deadline. Our proposed architecture (see Figure \ref{fig:TTCNN}) involves a set of shallow CNNs and a cloud CNN, where we transfer the first convolutional layers of the shallow CNNs to enhance the cloud CNN. 

\subsection{Model Architecture}
\label{modArc}
\subsubsection{Shallow CNNs}
\label{scnn}
Shallow CNNs make the proposed architecture more scalable since different shallow CNNs will contribute different knowledge to the cloud CNN. The cloud CNN absorbing different knowledge will have different generalization ability. 
To explore the designs of shallow CNNs, we consider the various design factors: CNN architecture, setup of hyper-parameters, training dataset, and the number of shallow CNNs.  

\begin{figure}[!htb]
  \centering
  \includegraphics[width= 0.35\textwidth]{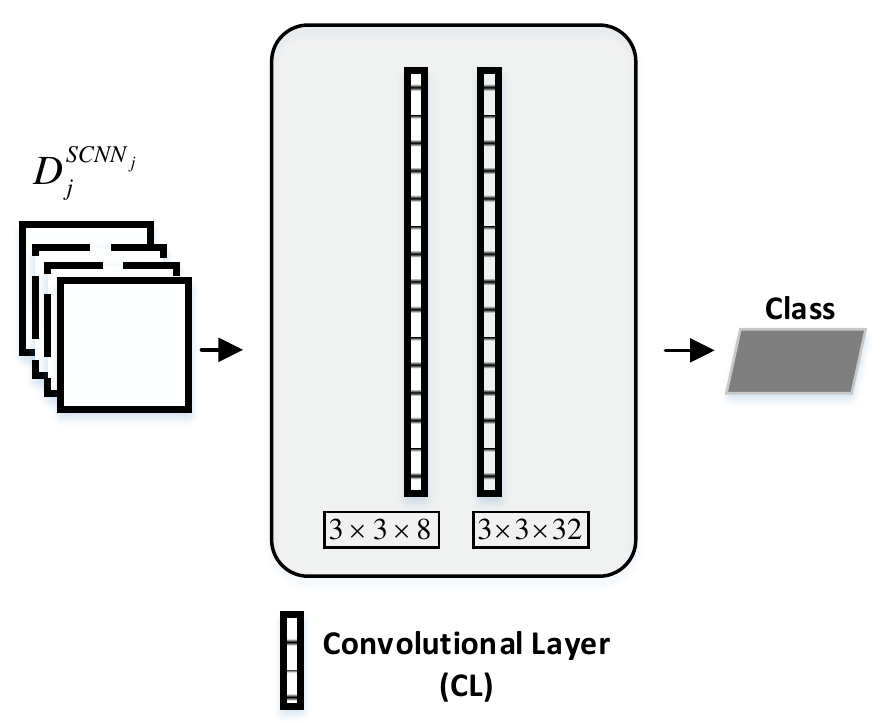}\\
  \caption{Architecture of shallow CNN for CIFAR-10. It contains two convolutional layers where the filter size is $3\times3$, and the numbers of filters are 8 and 32 for the first and second convolutional layers, respectively. }\label{fig:SCNN}
\end{figure}

\begin{figure}[!htb]
  \centering
  \includegraphics[width= 0.35\textwidth]{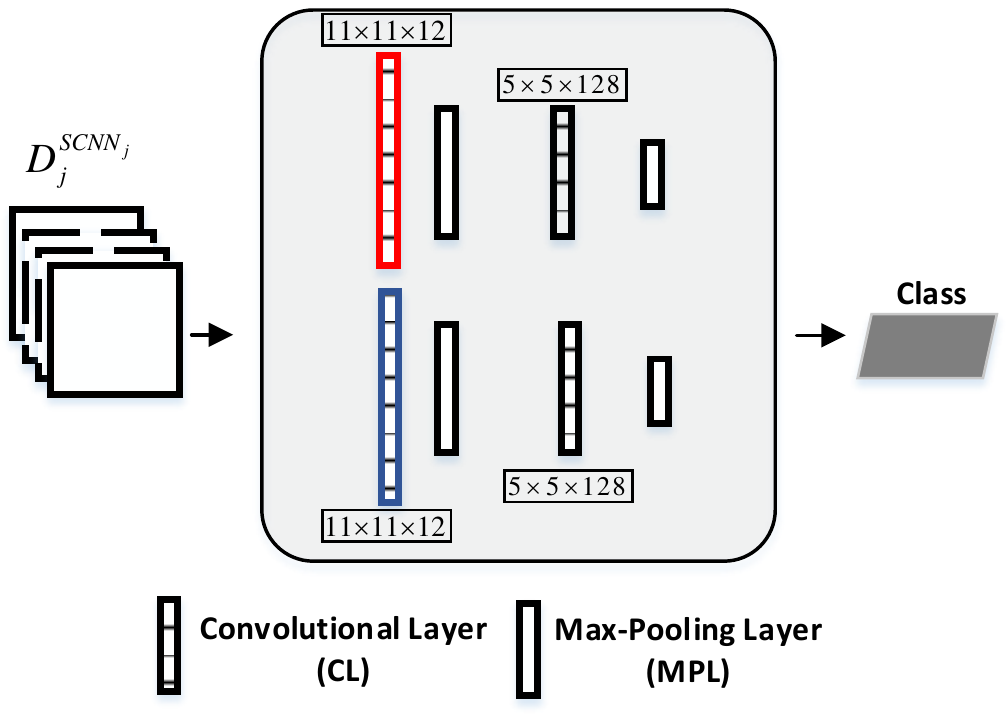}\\
  \caption{Architecture of shallow CNN for AlexNet. It contains two separate convolutional layers where the filter size of the first separate convolutional layers is $11\times11$, and the number of filters is 12.  }\label{fig:SCNNVGG}
\end{figure}

\textbf{Architecture.}  For shallow CNNs, we prefer the number of  layers small. Architecture of shallow CNNs shown in Figure \ref{fig:SCNN} is designed for CIFAR-10. The first convolutional layers in these shallow CNNs are transferred to initialize the first layer of the cloud CNN.
As shown in Figure \ref{fig:SCNN}, the shallow CNN contains only two convolutional layers for CIFAR-10. For the case of ImageNet the shallow CNN contains two separate parts to match the structure of the cloud CNN (AlexNet), see Figure \ref{fig:SCNNVGG}. 
In principle, we can design different architectures for different shallow CNNs. In this paper, we employ the same architecture for all the shallow CNNs for simplicity. 

\textbf{Hyper-parameters.} Different setup of hyper-parameters such as learning rate, learning max-iterations, and mini-batch for CNN would lead to different performance. However, selecting optimal hyper-parameters is still a challenge. For all the shallow CNNs, we can set different hyper-parameters to train these CNNs. In this paper, we set the same hyper-parameters for all the shallow CNNs.  The setting of the hyper-parameters for shallow CNNs in Figure \ref{fig:SCNN} are: learning rate: 0.01, mini-batch: 100, and learning max-iterations: 10,000. 
For shallow CNNs in Figure \ref{fig:SCNNVGG}, the hyper-parameters are: learning rate: 0.01, mini-batch: 256, and learning max-iterations: 10,000.

\textbf{Datasets.} Compared to transfer CNNs in~\cite{bansal2016pixelnet, shin2016deep, lee2017transfer}, the proposed architecture with multiple shallow CNNs allows us to extract knowledge efficiently from either  
identical or different datasets. Even if the distributions of the datasets are the same, we may still obtain different shallow CNNs by setting different hyper-parameters of the shallow CNNs. Moreover, when both the data distribution and the setup of hyper-parameters are identical for all shallow CNNs, we may still obtain different shallow CNNs since the training procedures based on stochastic gradient algorithm is stochastic in nature.  Furthermore, we can build datasets containing different image samples by following the same distribution, and the sizes of these datasets could be different as well. 

\textbf{Number of shallow CNNs.} Another key design factor is the number of shallow CNNs, and this is determined by how the first layer of the cloud CNN is decoupled. Different decouplings may lead to different performance of the cloud CNNs and may affect the difficulties of knowledge transfer. 

Combinations of the design factors discussed so far make the architecture of the proposed hierarchical transfer CNN more scalable. Since each factor plays a specific role, we may pay more attention to certain factors according to the requirements of the applications. 

\subsubsection{Cloud CNN}
\begin{figure}[!htb]
  \centering
  \includegraphics[width= 0.45\textwidth]{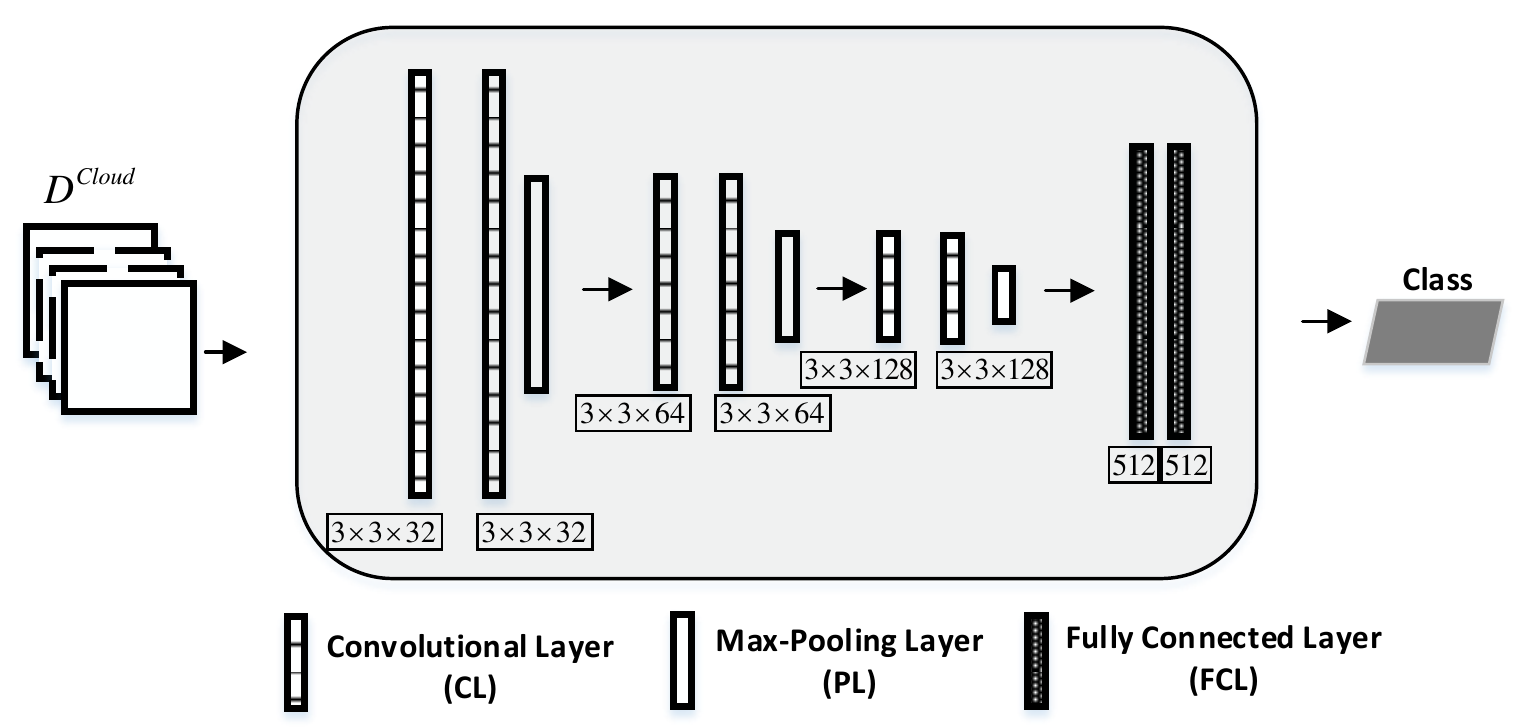}\\
  \caption{Architecture of the cloud CNN for CIFAR-10. It contains 6 convolutional layers, 3 max-pooling layers, and 2 fully connected layers. The convolutional layers share the same filter size: $3\times3$. The number of filters for 6 convolutional layers are 32, 32, 64, 64, 128, and 128. The number of neurons in the fully connected layers is 512. }\label{fig:CCNN}
\end{figure}

This paper considers two cloud CNNs for CIFAR-10 and ImageNet, respectively. Figure \ref{fig:CCNN} illustrates the architecture of the cloud CNN for CIFAR-10. It consists of 6 convolutional layers interspersed with 3 max pooling layers. 
This architecture is much shallower than that of the normal cloud CNN. Therefore, we can examine the efficiency of the proposed architecture in different situations. 
We construct this architecture by following the common ConvNet architecture\footnote{http://cs231n.github.io/convolutional-networks/}  INPUT $\to$ [[CONV $\to$ RELU]*N $\to$ POOL?]*M $\to$ [FC $\to$ RELU]*K $\to$ FC, where multiple convolution layers followed by a pooling layer form a complex layer, and then fully connected layers are added to generate the output. 
In the case of ImageNet, AlexNet is employed as the cloud CNN due to its state-of-the-art performance on image classification \cite{NIPS2012_0534, russakovsky2015imagenet}.

\begin{figure*}[!htb]
  \centering
  \includegraphics[width= 0.95\textwidth]{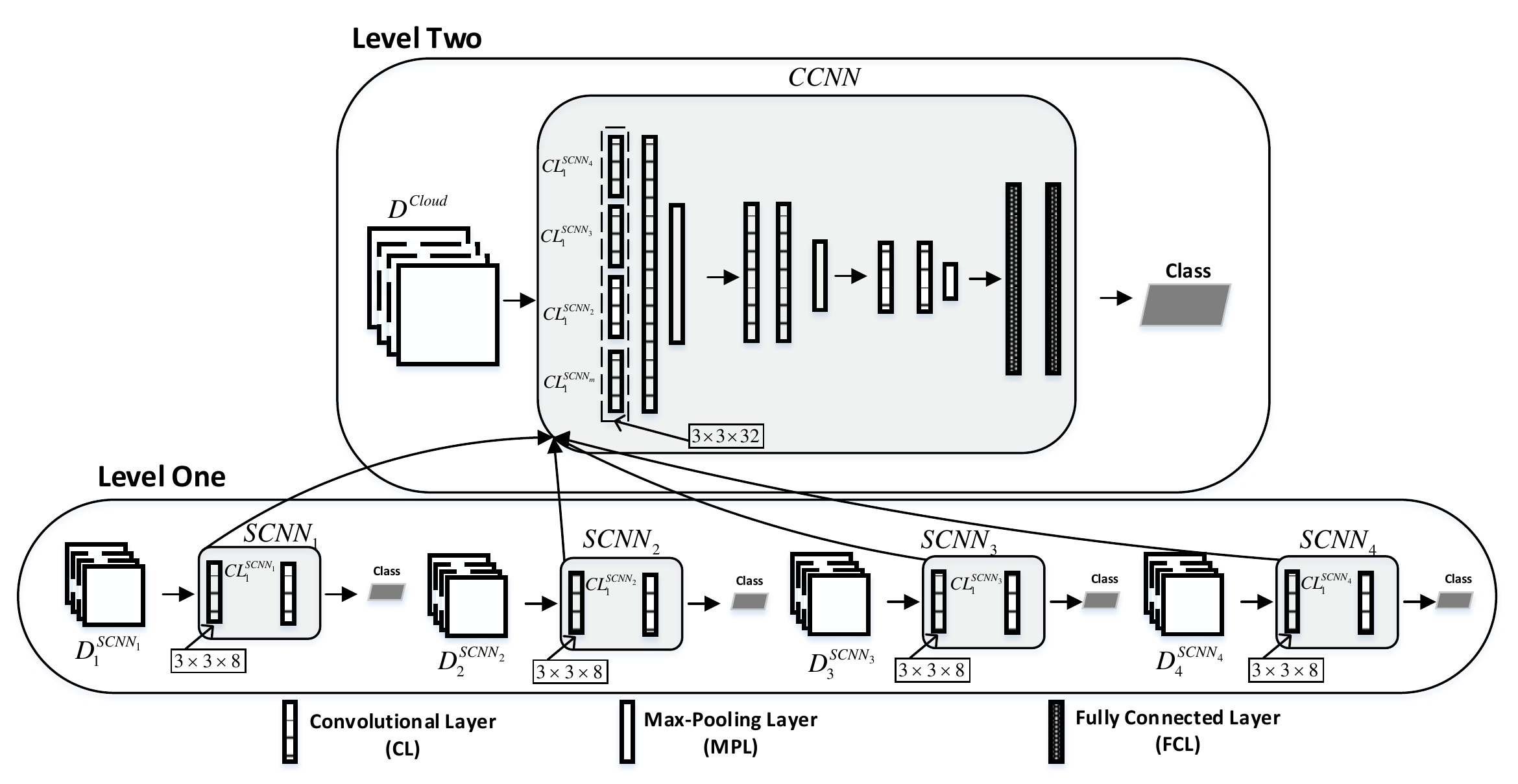}\\
  \caption{Architecture of HTCNN for CIFAR-10. This architecture consists of 4 shallow CNNs and a cloud CNN. The setup of first convolutional layer in these shallow CNNs is $3\times3\times8$, where the filter size is $3\times3$, and the number of filters is $8$. For the cloud CNN, the first convolutional layer  is $3\times3\times32$.}\label{fig:TLTCNN}
\end{figure*}

\subsubsection{Transfer Learning}
After completing the architecture design of shallow CNNs and cloud CNN, we will decide which parts of these shallow CNNs are used to transfer knowledge to enhance the cloud CNN. Yosinski \etal \cite{yosinski2014transferable} analyze the performance differences when transferring different numbers of CNN layers and show that shallow layers such as the first layer will be generally beneficial for constructing transfer CNN. Moreover, both fine-tuning and frozen transferred layers could be useful to improve CNN performance. In this paper, we prefer the fine-tuning method rather than the frozen method. 

In the case of HTCNN for CIFAR-10, we first divide the first layer ($3\times3\times32$) of the  cloud CNN into four parts as shown in Figure \ref{fig:TLTCNN}. Each part shares the same setup $3\times3\times8$, where the filter size is $3\times3$, and the number of filters is $8$. We train four shallow CNNs on their own dataset and transfer their first layers to initialize the four parts of the first layer of the cloud CNN.

\begin{figure*}[!htb]
  \centering
  \includegraphics[width= 0.95\textwidth]{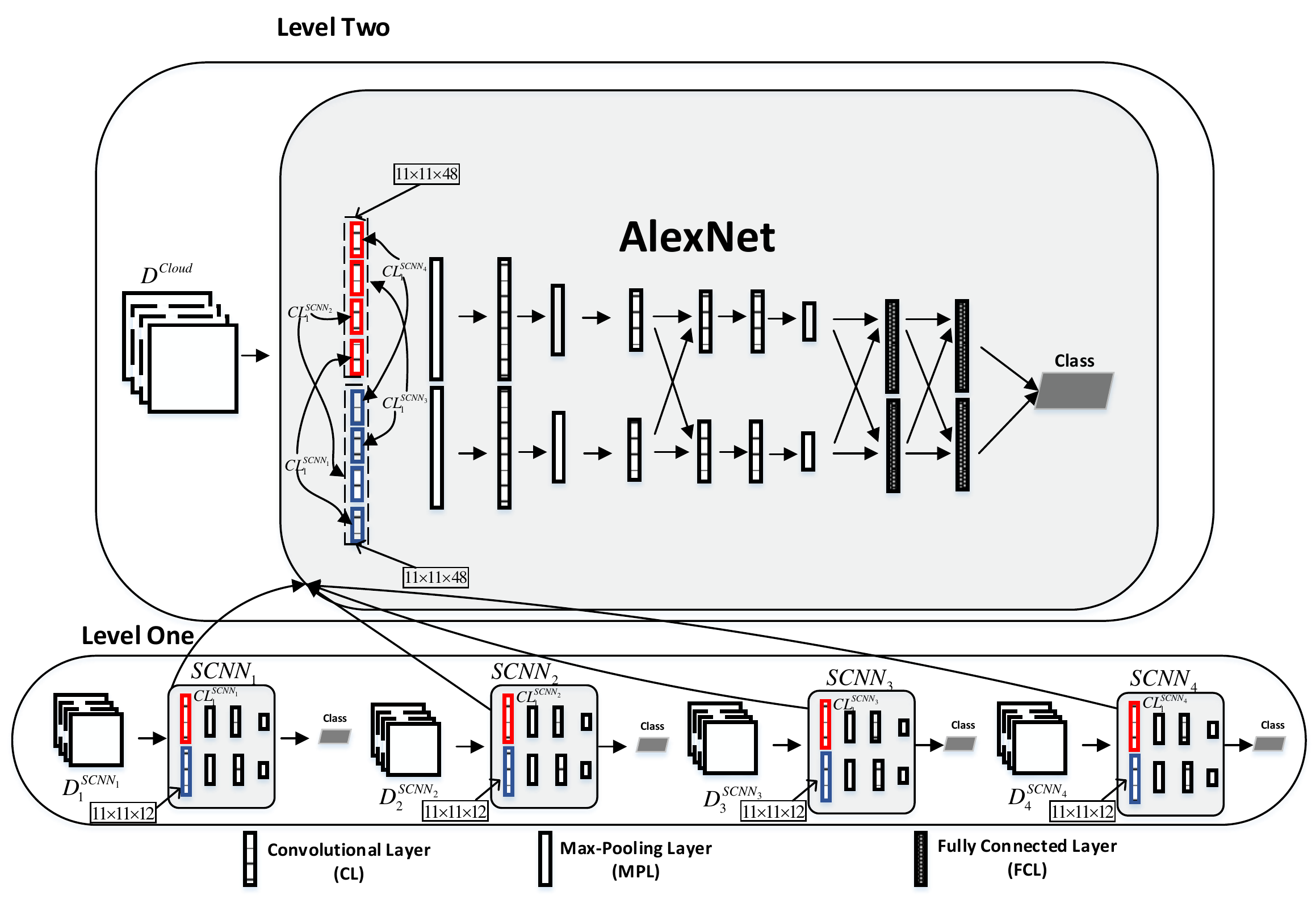}\\
  \caption{Architecture of enhancing AlexNet in the framework of HTCNN. This architecture consists of 4 shallow CNNs. The setup of first convolutional layer in these shallow CNNs is $11\times11\times12$, where the filter size is $11\times11$, and the number of filters is $12$. For AlexNet,  the first convolutional layer has two $11\times11\times48$ parts. We transfer the first convolutional layer of  4 shallow CNNs to initialize the first convolutional layer of AlexNet, with upper and lower half of the layer go to the corresponding locations in AlexNet, as color coded here.}\label{fig:TLTAlexNet}
\end{figure*}

In the case of enhancing AlexNet in the framework of HTCNN, as shown in Figure \ref{fig:TLTAlexNet}, we first divide the first layer ($11\times11\times48$) of AlexNet into four parts. Each part shares the same setup $11\times11\times12$, where the filter size is $11\times11$, and the number of filters is $12$. We train four shallow CNNs on their own dataset and transfer their first layer to initialize the first layer in AlexNet.

\subsection{Loss Function, Training, and Optimization}
\label{trainopti}

We use softmax cross entropy as the loss function for both the cloud CNN and shallow CNNs.
We train the full model end-to-end in a single step of optimization. The shallow CNNs are initialized randomly. While the first layer of the cloud CNN is initialized by injecting the first layers of shallow CNNs, other layers are initialized with random weights. We use stochastic gradient descent with momentum 0.9 to train the weights of the cloud CNNs. 

The proposed method could be analyzed using the framework of deep transfer learning~\cite{TheoryTransferLearningGalanti2016}.
Here we have a transfer learning setting $\mathcal{T} = (T, \mathcal{L}, \mathcal{E})$ being a realizable classification or regression setting. $T$ specifies a learning setting $ T = (H, Z, l)$ where $H$ is the hypothesis set, $ Z = X \times Y$ is an example set, $X$ is the instance set and $Y$ is the label set, and $l$ is a loss function. $\mathcal{L}$ is a hypothesis class family.
There are $k$ shallow CNNs for source tasks $d_1, d_2, \cdots, d_k \in \mathcal{E} $ and one cloud CNN for a target task $d_t \in \mathcal{E} $ in the environment $\mathcal{E}$.  

Suppose the samples are generated by a process $\mathcal{D}[k,p,q]$, that samples $ i.i.d.$ $k+1$ tasks $d_1, d_2, \cdots,~d_k,~d_t $ from a distribution $D$ and returns samples $S=(s[1,k],s_t)$ that contains $k$ source data sets $s_i \sim d_i^p$,  and one target data set $s_t \sim d_t^q$, where $p$ is the size of source dataset $d_i$, and $q$ is the size of target dataset $d_t$. Then the proposed method, $g(S)$ is a narrowing process: the $k$ shallow CNNs intend to obtain coarse grained models that help narrow down the hypothesis in the hypothesis set. When the part of the obtained shallow CNNs are injected into the more sophisticated cloud CNN,  $g(S)$ plays the role of a simplifier:
\begin{equation}
P_{S \sim D}  \left[ \epsilon_{d_t} (g(S)) \le \inf_{c \in H} \epsilon_{d_t}(c)  + \epsilon  \right] \ge 1- \delta \; , 
\forall \epsilon, \delta
\end{equation}
where $\epsilon_{d_t} (g(S))$ is the generalization error after knowledge transfer.
In principle, it also can be shown that (1) the number of shallow CNNs, $k$ should increase to improve performance if the amount of source data samples increase; (2)  $k$ should increase if the amount of target data samples decrease, since more transferred knowledge is needed to reduce the search space of the hypothesis set for the target task.

\section{Experiment}
\label{sec5}

Here we evaluate our proposed method by completing image classification on CIFAR-10 and ImageNet. Specifically, we validate our model with two image classification tasks:
\begin{itemize}
\item Firstly, we evaluate our proposed method by completing image classification on CIFAR-10 and the effect of different settings such as the datasets used (identical or localized), dropout, and number of shallow CNNs are examined.
\item Secondly, we validate our approach through transferring multi-source knowledge to enhance AlexNet for image classification on ImageNet.
\end{itemize}

In both cases, we train the proposed hierarchical transfer CNN (HTCNN) and plain cloud CNN (CCNN) up to a certain number of epoch, and test the trained HTCNN and CCNN with testing datasets to obtain the testing accuracy.
Then we compare the generalization performance of HTCNN with CCNN based on their testing accuracy. 
We repeat this with various number of epoch.

\subsection{Evaluation Metrics}
To evaluate the speedup performance of the proposed method, we design two metrics: Average Accuracy Gain (AAG) and Percentage of Better Performance (PBP). 
AAG is defined by
\begin{equation}
AAG=\frac{\sum_{i=1}^{N_{iter}} Acc^{HTCNN}_{i} - Acc^{CCNN}_{i}}{N_{iter}} 
\label{equa:aag}
\end{equation}
where $N_{iter}$ is the max-iterations, $Acc^{HTCNN}_{i}$ and $Acc^{CCNN}_{i}$ are the \emph{testing accuracy} of the proposed hierarchical transfer CNN (HTCNN) and cloud CNN only (CCNN) obtained after the $i$th epoch ($\times$1000 iterations), respectively. 

An indicator function $ BP_{i}$ is defined as
\begin{equation}
BP_{i}=\left\{
\begin{aligned}
1 &&&&& if Acc^{HTCNN}_{i} > Acc^{CCNN}_{i}\\
0 &&&&&  otherwise \\
\end{aligned}
\right.
\label{equa:bp}
\end{equation}
which returns 1 whenever the HTCNN outperforms the CCNN.
Then, PBP is given by
\begin{equation}
PBP=\frac{\sum_{i=1}^{N_{iter}} BP_{i}}{N_{iter}} .
\label{equa:bpp}
\end{equation}
PBP counts the percentage of occurrences that HTCNN outperforms CCNN for models obtained after various number of iterations. 

\subsection{Experiments on CIFAR-10}

\subsubsection{CIFAR-10}

The CIFAR-10 dataset \footnote{https://www.cs.toronto.edu/~kriz/cifar.html} consists of 60,000 32x32 RGB images in 10 classes with 6,000 images per class. It includes 50,000 training images and 10,000 test images distributed in six datasets, where each dataset for training shares similar distributions of categories indicated as in Table \ref{tab:cifar}. As shown in Table \ref{tab:cifar}, although numbers of different categories of images in different datasets are not the same, the proportions of images distributed among these categories are similar.

\begin{table*}[htb]\normalsize
\caption{Data distribution of training datasets of CIFAR-10}
    \begin{center}
    \begin{tabular}{l|c|c|c|c|c|c|c|c|c|c}
    	\hline
    	\textbf{Category}	& 	Airplane 	& 	Automobile 	&	Bird 	&	 Cat 	&	 Deer 	&	 Dog & Frog  	&	Horse  	&	Ship   	&	Truck \\
    	\hline
		Dataset 1 &	1005   	&	974  	&	1032  	&	1016   	&	999   	&	937  	&	1030  	&	1001  	&	1025  	& 	981 \\
		Dataset 2 &	984  	&	1007  	&	1010  	& 	995  	&	1010  	& 	988  	&	1008  	&	1026  	& 	987   	&	985 \\
		Dataset 3 &	994  	&	1042  	& 	965  	& 	997   	&	990  	&	1029  	& 	978  	&	1015 	&  	961  	&	1029 \\
		Dataset 4 &	1003  	& 	963  	&	1041   	&	976  	&	1004  	&	1021  	&	1004  	& 	981  	&	1024 	&  	983 \\
		Dataset 5 &	1014  	&	1014  	& 	952  	&	1016  	& 	997  	&	1025   	&	980   	&	977  	&	1003 	& 	1022 \\
		\hline
    \end{tabular}
    \end{center}
\label{tab:cifar}
\end{table*}

\subsubsection{Experiment Setup for CIFAR-10}
The experiment contains two cases: identical data and data locality. We set the number of the shallow CNNs as four. In the case of identical data, we use Dataset1 to Dataset5 for training all four shallow CNNs (SCNNs) and the cloud CNN. To address data locality in the second case, we use Dataset1 shown in Table \ref{tab:cifar} to train the CCNN, while use Dataset2 to train SCNN1, Dataset3 to train SCNN2, etc. as indicated in Figure \ref{fig:TLTCNN}. Our implementation uses TensorFlow \cite{abadi2016tensorflow}. Training the cloud CNN for CIFAR-10 on a Titan X GPU takes about four hours to converge.

The models have been trained and tested nine times and the bar plots are given to show the average generalization performance together with the maximum and minimum testing accuracy as head and foot along the bar.

\subsubsection{Experimental results on CIFAR-10}
\begin{figure}[!htb]
  \centering
  \includegraphics[width= 0.5\textwidth]{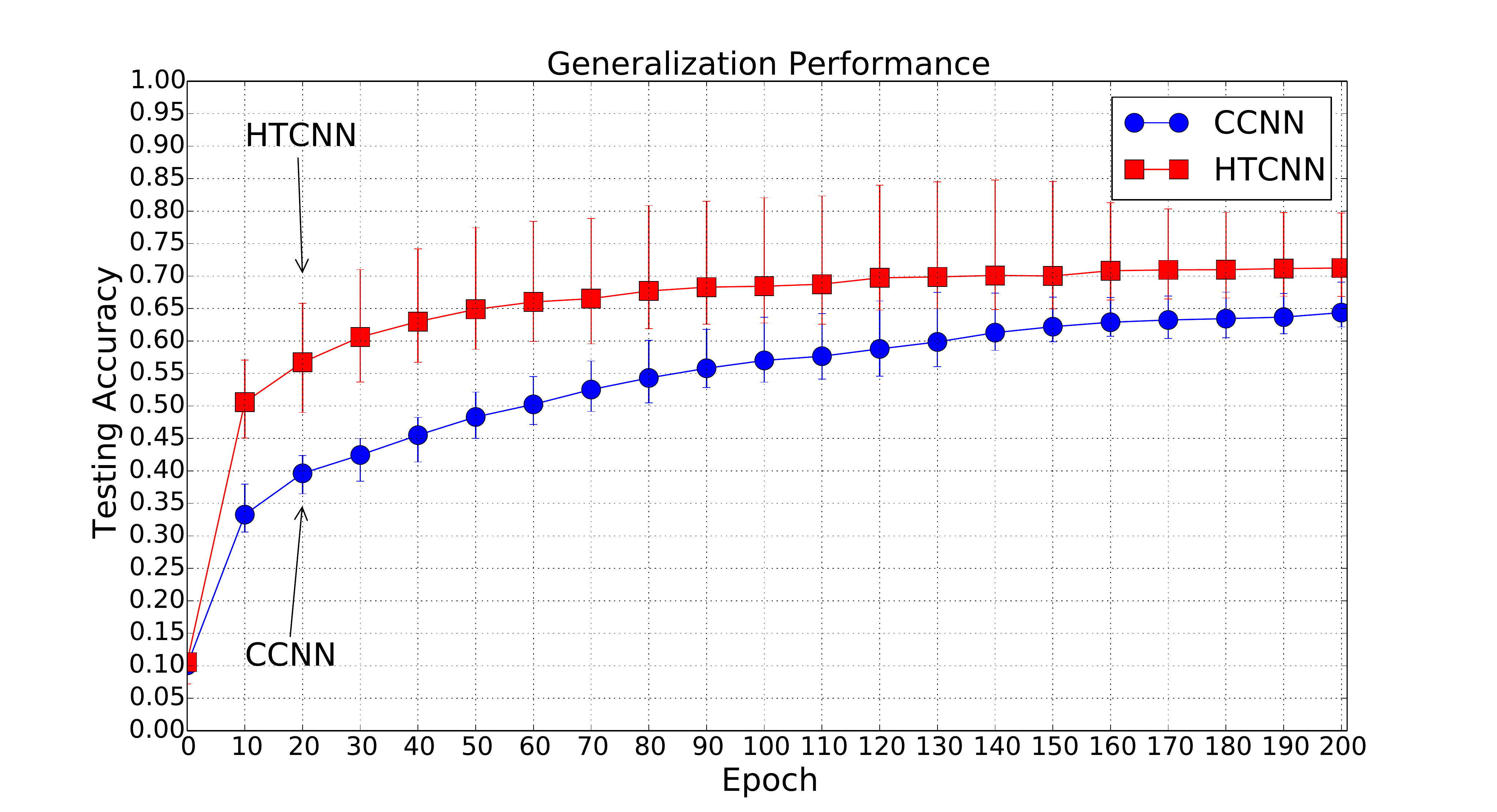}\\
  \caption{Comparing generalization performance. x-axis is the training epoch. Each epoch contains 1,000 iterations. y-axis is the testing accuracy.}\label{fig:cifar-identity}
\end{figure}

\textbf{Identical Data. } We build a hierarchical transfer CNN as shown in Figure \ref{fig:TLTCNN} and use the cloud CNN (see Figure \ref{fig:CCNN}) as baseline. The generalization performance are shown in Figure \ref{fig:cifar-identity}. Comparing models obtained after short training period (from Epoch 0 to Epoch 60), intermediate period (from Epoch 60 to Epoch 140), and long period (from Epoch 140 to Epoch 200), the testing accuracies of HTCNN has significant gain over 
those of CCNN, especially when the models are obtained after short training period. 
This indicates that HTCNN could provide better performance than CCNN when applied to real-time applications.
For instance, if the deadline is 60 epoch, the HTCNN will have $15\%$ gain in model accuracy compared to CCNN.
In terms of AAG and PBP, we obtain AAG=0.12 and PBP=1.0, which indicate a $12\%$ gain in model accuracy on average and HTCNN outperforms CCNN across the board.

\begin{figure}[!htb]
  \centering
  \includegraphics[width= 0.5\textwidth]{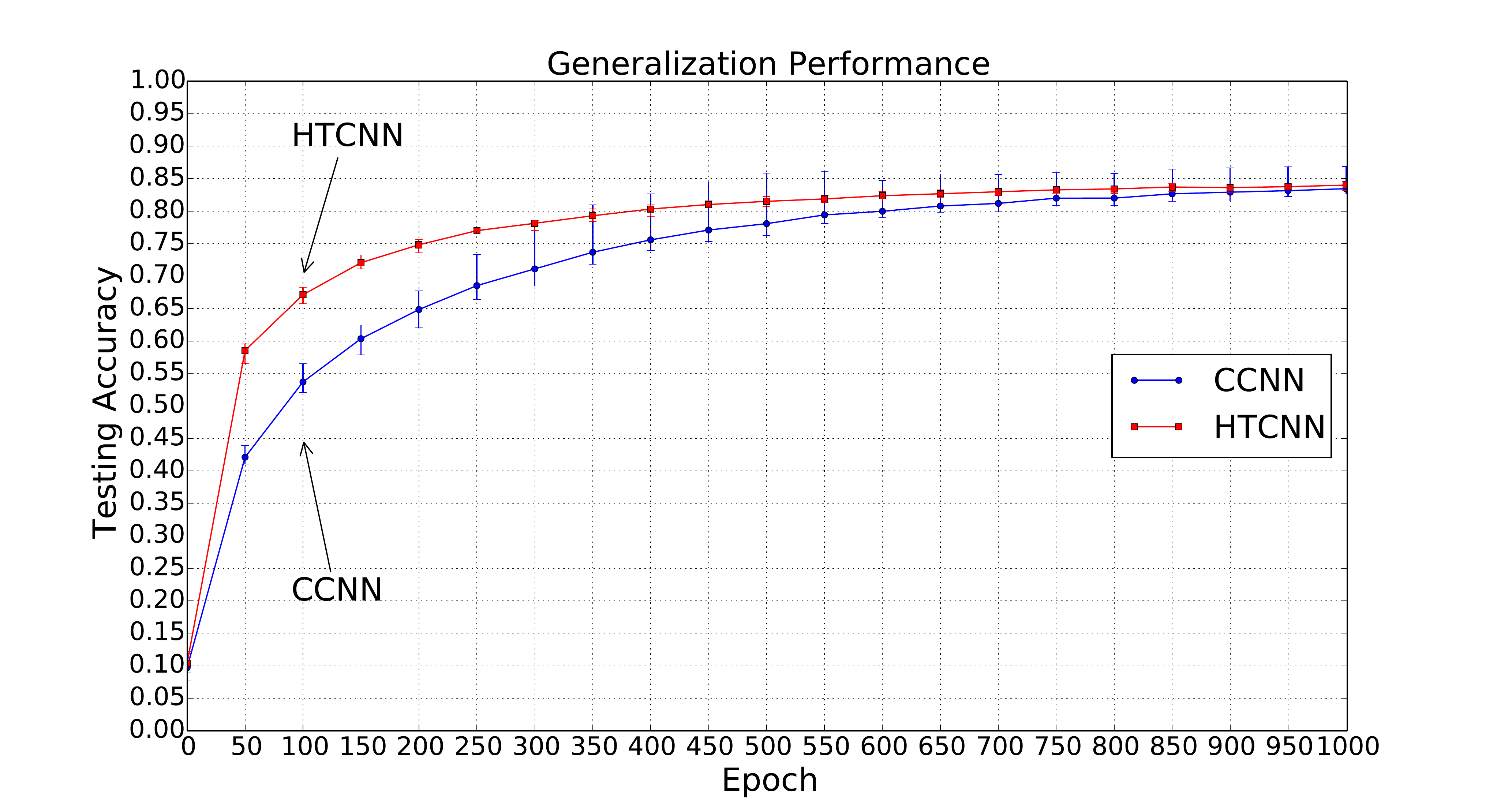}\\
  \caption{Comparing generalization performance when applying Dropout to CCNN and HTCNN.  }\label{fig:cifar-dropout}
\end{figure}

Dropout \cite{srivastava2014dropout} is a simple way to prevent neural networks from overfitting. Therefore, we examine this technique by applying dropout to the cloud CNN. Specifically, we apply dropout to convolutional layers and fully connected layers by setting dropout probabilities as 0.8 and 0.5, respectively. As shown in Figure \ref{fig:cifar-dropout}, 
dropout slows down the convergence of the model but improves the model accuracy compared to that without dropout as expected. Unlike the case without dropout, the speedup effect is only significant during the early learning stage (from Epoch 0 to Epoch 300) and vanished in the converged stage (from Epoch 900 to Epoch 1,000). It implies that CCNN with dropout is able to achieve similar generalization performance of HTCNN after convergence. 

\begin{figure}[!htb]
  \centering
  \includegraphics[width= 0.5\textwidth]{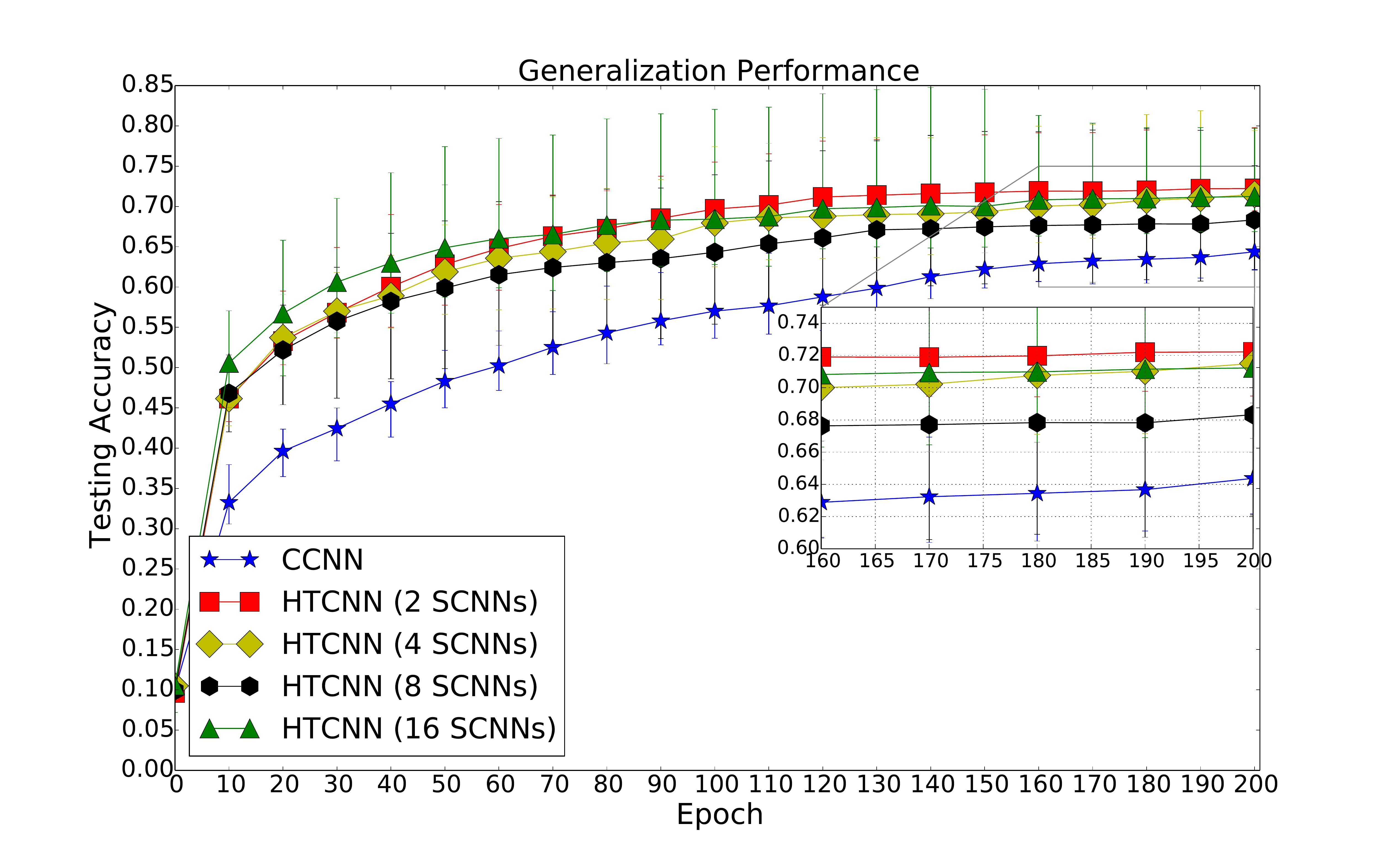}\\
  \caption{Comparing generalization performance when applying different numbers of SCNNs to build HTCNN.}\label{fig:cifar-compare}
\end{figure}

In section \ref{scnn}, the number of SCNNs is also a key factor that will affect model performance. Therefore, we examine this factor by using different number of SCNNs, namely 2, 4, 8, and 16 SCNNs to initialize the first layer of the cloud CNN to build HTCNN without dropout, while keeping the size of the first layer of the cloud CNN constant. The results are given in Figure \ref{fig:cifar-compare}. All HTCNNs with different number of SCNNs outperform CCNN, especially in the early learning stage. It seems that HTCNN with more shallow CNNs (the case of 16 SCNNs) outperform other models in the early training stage. In the middle learning stage, testing accuracies of these HTCNNs are similar, while in the converged stage, HTCNN with 8 SCNNs is worse than other HTCNNs, and the HTCNNs with 2, 4, and 16 SCNNs converge to similar accuracies. 

\begin{figure}[!htb]
  \centering
  \includegraphics[width= 0.5\textwidth]{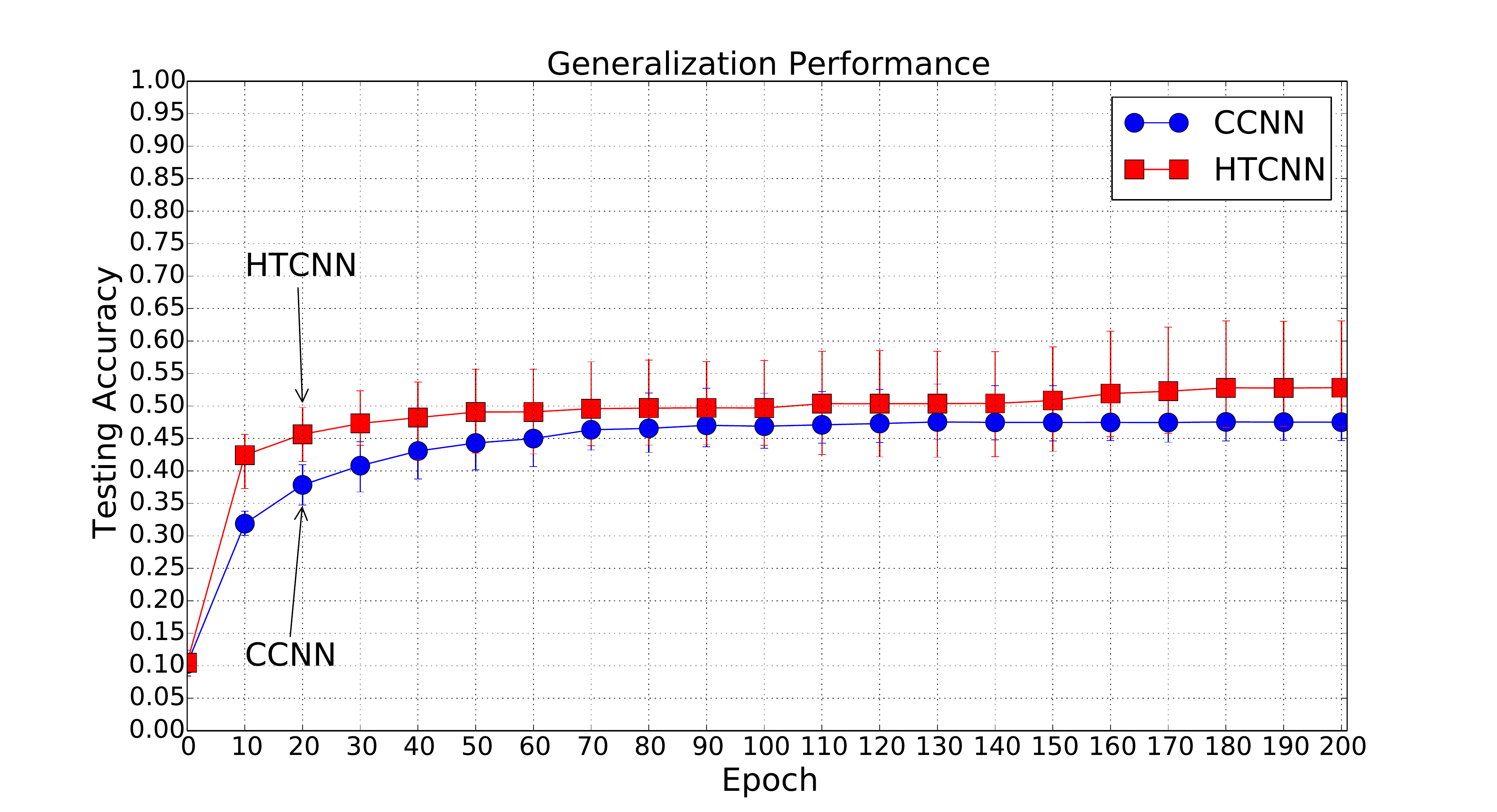}\\
  \caption{Comparing generalization performance when training sets for shallow CNNs and cloud CNN are not identical.}\label{fig:cifar-locality}
\end{figure}

\textbf{Data Locality. } We examine the generalization performance of HTCNN without dropout when applying different datasets to train different SCNNs. As shown in Figure \ref{fig:cifar-locality}, HTCNN outperforms CCNN in the entire learning processes. Moreover, in the early stage (from Epoch 0 to Epoch 60), HTCNN performs better than CCNN, which is consistent with the case of identical data. However, compared to the case of identical data with AAG=0.12, the accuracy gains are less with AAG=0.04. 

\subsection{Experiments on ImageNet}

\subsubsection{ImageNet}

We down-sampled the images to a fixed resolution of $256 \times 256$ in order to meet a constant input dimensionality by cropping out the central patch from the resulting image. And we subtract the mean activity over the training set from each pixel.

\subsubsection{Experiment Setup for ImageNet}
We design two groups of experiments to verify the efficiency of the proposed model. Class identity is the first group where we segment the ImageNet into two datasets $D_{1}$ and $D_{2}$ including 640,610 and 640,557 images, respectively. Both of these two datasets contains 1,000 classes. We employ $D_{1}$ to train cloud AlexNet while $D_{2}$ for training small CNN. Class locality is the other group where  we segment the ImageNet into two parts with each part containing 1,000 classes. The first part including 640,610 images distributed into 1,000 classes is used for training AlexNet, while the other part containing 640,557 images is used to train four shallow CNNs. Then we transfer the first layer of these four shallow CNNs to initialize the first layer of AlexNet to build HTAlexNet. 

\subsubsection{Experimental results on ImageNet}

\begin{figure}
	\centering
	\includegraphics[width= 90mm]{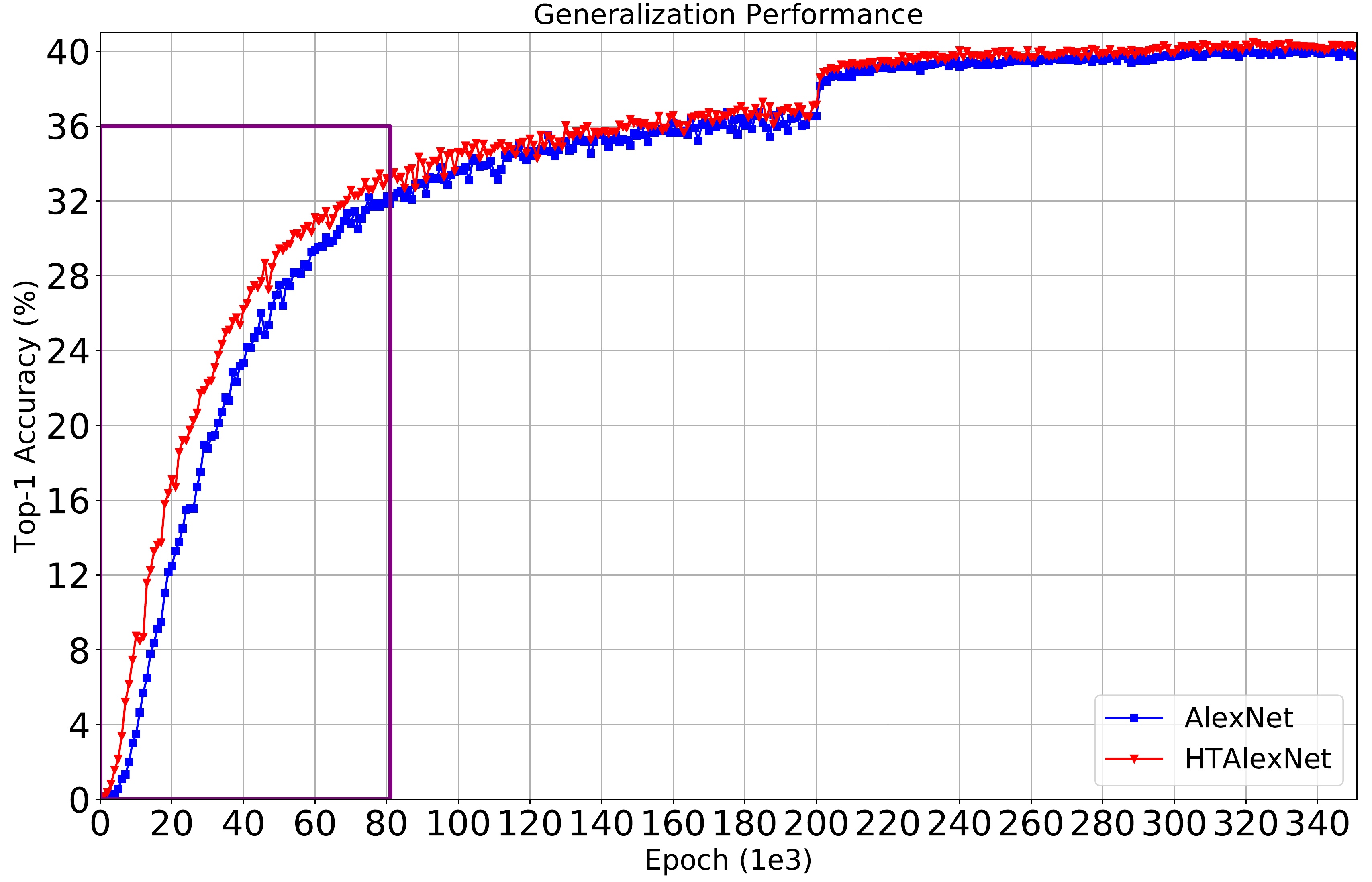}\\  (a)  \\
	\includegraphics[width= 90mm]{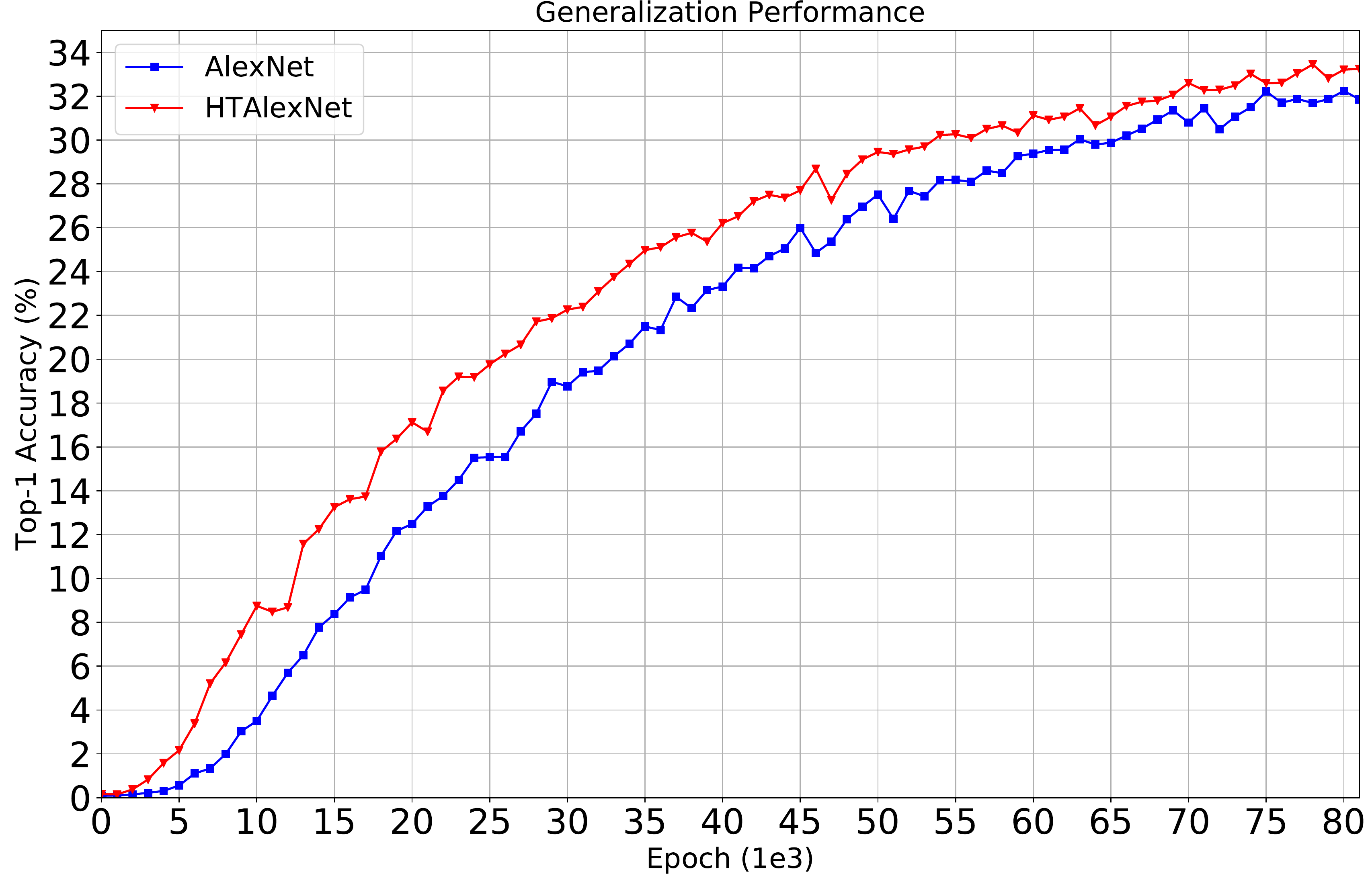}\\  (b) \\
	\caption{(a) Comparison of Top-1 generalization performance between AlexNet and the proposed hierarchical transfer AlexNet (HTAlexNet). The details of the first 80 epoch is given in (b). }\label{fig:imagenet-locality}
\end{figure}

We examine our model by comparing generalization performance of hierarchical  transfer AlexNet (HTAlexNet) and AlexNet by training these two models on ImageNet. Figure \ref{fig:imagenet-locality} shows about 5\% Top-1 accuracy gain at Epoch 20. 
These results demonstrate that the proposed method would be useful for practitioners in real-world application-oriented environments. Additionally, the performance differences in this case are much smaller that those in the case of CIFAR-10 for two reasons. One is that the shallow CNN are not able to provide efficient transfer features to enhance cloud AlexNet. The other is that AlexNet have more filters to make itself more stable than the cloud model for CIFAR-10.


In all of our experiments, it is observed that the generalization performance such as model accuracy of the HTCNN is better than that of the CCNN after model convergence for the CIFAR-10 case. This is consistent with the results in \cite{sharif2014cnn, girshick2016region, huh2016makes, lee2017transfer}. For the ImageNet case, in the early training stage, we still observe the accuracy gains even if we cannot obtain the higher converged performances with AlexNet.


\section{Conclusion and Future Work}
\label{sec7}

In this paper, we propose a novel hierarchical transfer CNN for image classification by transferring knowledge from multiple data sources. This architecture consists of a group of shallow CNNs and a cloud CNN. After the training of the shallow CNNs is complete, the first layers of these shallow CNNs are extracted to initialize the first layer of the cloud CNN. 
The proposed method is evaluated using CIFAR-10 and ImageNet datasets. Experimental results demonstrate the proposed method could improve the generalization performance of CNN under various settings. It could also boost the  generalization performance of CNN during the early stage of learning. This makes the proposed method very attractive for real-time applications where a satisfactory image classifier needs to be trained within a deadline.

\section*{Acknowledgment}
This research work is supported in part by the U.S. Office of the Assistant Secretary of Defense for Research and
Engineering (OASD(R\&E)) under agreement number FA8750-15-2-0119. The U.S.
Government is authorized to reproduce and
distribute reprints for Governmental purposes notwithstanding
any copyright notation thereon. The views and conclusions
contained herein are those of the authors and should not be
interpreted as necessarily representing the official policies or
endorsements, either expressed or implied, of the U.S. Office of
the Assistant Secretary of Defense for Research and Engineering
(OASD(R\&E)) or the U.S. Government.



\bibliographystyle{IEEEtran}
\bibliography{References}
%



\end{document}